\documentclass{article} 
\usepackage{iclr2025_conference,times}

\usepackage{wrapfig}
\usepackage{lipsum} 
\usepackage{wrapfig}

\usepackage{amsmath,amsfonts,bm}
\usepackage{multicol}

\def\eqref#1{equation~\ref{#1}}

\def\1{\bm{1}}

\DeclareMathAlphabet{\mathsfit}{\encodingdefault}{\sfdefault}{m}{sl}
\SetMathAlphabet{\mathsfit}{bold}{\encodingdefault}{\sfdefault}{bx}{n}

\usepackage{booktabs}      
\usepackage{multirow}      
\usepackage{amssymb}       
\usepackage{graphicx, subfigure}     

\usepackage{colortbl}
\usepackage{xcolor}

\usepackage{hyperref}
\usepackage{url}
\usepackage{multicol}
\usepackage{aliascnt}
\usepackage{adjustbox}
\usepackage{siunitx} 
\usepackage{booktabs}
\usepackage{multirow} 
\usepackage{enumitem}
\usepackage{booktabs} 
\usepackage{amssymb}  
\usepackage{amsmath}  
\usepackage{graphicx} 
\usepackage{longtable} 
\usepackage{graphicx}
\usepackage{subcaption}
\usepackage{calc}
\usepackage[most]{tcolorbox}
\usepackage{listings}

\definecolor{uclablue}{rgb}{0.15, 0.45, 0.68}
\hypersetup{
    breaklinks,
    citecolor=uclablue,
    colorlinks=true,
}

\newtcolorbox{AIbox}[2][]{aibox,title=#2,#1}
\tcbset{
  aibox/.style={
    top=10pt,
    colframe=black,
    colbacktitle=black,
    coltitle=white,
    center,
  }
}

\lstdefinelanguage{prompt}{
    basicstyle=\scriptsize\ttfamily, 
    mathescape=true,        
    escapebegin=\color{latentcolor},  
    escapeend={},
    escapechar=@,
    stringstyle = \color{myorange},
    showstringspaces = false,
    moredelim = [s][\color{mypink}]{`}{`},
    moredelim = [s][\color{mybrown}]{```json}{```},
    moredelim = [s][\color{latentcolor}]{<StartOfLatent>}{<EndOfLatent>},
    literate = %
        {\ \ a.\ }{{\textcolor{mypurple}{\ \ a.\ }}}5
        {\ \ b.\ }{{\textcolor{mypurple}{\ \ b.\ }}}5
        {\ \ c.\ }{{\textcolor{mypurple}{\ \ c.\ }}}5
        {\ \ d.\ }{{\textcolor{mypurple}{\ \ d.\ }}}5
        {\ \ e.\ }{{\textcolor{mypurple}{\ \ e.\ }}}5
        {\ \ f.\ }{{\textcolor{mypurple}{\ \ f.\ }}}5
        {\ \ g.\ }{{\textcolor{mypurple}{\ \ g.\ }}}5
        {\ \ h.\ }{{\textcolor{mypurple}{\ \ h.\ }}}5
        {\ I.\ }{{\textcolor{mypurple}{\ I.\ }}}4
        {\ II.\ }{{\textcolor{mypurple}{\ II.\ }}}5
        {\ III.\ }{{\textcolor{mypurple}{\ III.\ }}}6
        {\ IV.\ }{{\textcolor{mypurple}{\ IV.\ }}}5
        {\ V.\ }{{\textcolor{mypurple}{\ V.\ }}}4
}

\newtcblisting[auto counter, number within=subsection]{prompt}[4][]{%
  before skip=6pt, after skip=6pt,    
  top=6pt, bottom=6pt, left=7pt, right=7pt, 
  enhanced, breakable,
  colback=white, colframe=gray!65,
  boxrule=0.6pt, arc=2pt,
  drop shadow={black!15},             
  listing only,
  listing options={
    language=prompt,
    upquote=true,
    basicstyle=\scriptsize\ttfamily \setlength{\baselineskip}{1.1\baselineskip},
    columns=fullflexible,
    breaklines=true,
    breakindent=0pt,
    xleftmargin=\ifx\empty#2\empty0pt\else#2\fi,
    xrightmargin=\ifx\empty#3\empty0pt\else#3\fi,
    aboveskip=0pt,                    
    belowskip=0pt,                    
    showstringspaces=false,
  },
  title={\ifstrempty{#4}{Prompt \thetcbcounter}{Prompt \thetcbcounter: #4}},
  colbacktitle=gray!6, coltitle=black,
  attach boxed title to top left={yshift=-1mm, xshift=6pt},
  boxed title style={colback=gray!6, boxrule=0pt, sharp corners},
  #1
}

\newtcblisting[auto counter, number within=subsection]{showcase}[4][]{%
  before=\par\vspace{\baselineskip},
  after=\par,
  width=\linewidth,
  enhanced,
  arc=0em,
  boxrule=1pt,
  listing only,
  listing options={
    language=prompt,
    upquote=true,
    basicstyle=\scriptsize\ttfamily \setlength{\baselineskip}{1.1\baselineskip},
    breaklines=true,
    breakindent=0pt,
    xleftmargin=\ifx\empty#2\empty-12pt\else#2\fi,
    xrightmargin=\ifx\empty#3\empty-5pt\else#3\fi,
    aboveskip=-4pt,
    belowskip=-4pt,
    columns=fullflexible,
    },
  colback=white,
  colframe=gray,
  colbacktitle=gray!5,
  coltitle=black,
  attach boxed title to top center={yshift=-3mm},
  box align=center,
  parbox=false,
  title={\ifstrempty{#4}{Example \thetcbcounter}{Example \thetcbcounter: #4}},
  #1
}

\definecolor{linkColor}{rgb}{0.2,0.4,0.6}
\definecolor{myblue}{HTML}{0379AC}
\definecolor{myred}{HTML}{A50E50}
\definecolor{myorange}{RGB}{238, 133, 74}
\definecolor{latentcolor}{named}{cyan}
\definecolor{normalcolor}{RGB}{0, 0, 0}
\usepackage{marvosym}

\newcommand{\ours}{RLPT}

\title{Reinforcement Learning on Pre-Training Data}

\author{
Siheng Li$^{1,3,}$\thanks{\ Work completed during an internship at Tencent.}~~$^{,}$\thanks{\ The first three authors contributed equally to this work.}~~, Kejiao Li$^{1,\dagger}$, Zenan Xu$^{1,\dagger}$, Guanhua Huang$^1$, Evander Yang$^1$, Kun Li$^{1,3,*}$,\\ 
Haoyuan Wu$^{1}$, Jiajia Wu$^{1}$, Zihao Zheng$^{1}$, Chenchen Zhang$^{1}$, Kun Shi$^{1}$, Kyrierl Deng$^{1}$, Qi Yi$^{1}$,\\ 
Ruibin Xiong$^{1}$, Tingqiang Xu$^{1,*}$, Yuhao Jiang$^{1}$, Jianfeng Yan$^{1}$, Yuyuan Zeng$^{1}$, Guanghui Xu$^{1}$,\\ 
Jinbao Xue$^{2}$, Zhijiang Xu$^{2}$, Zheng Fang$^{2}$, Shuai Li$^{2}$, Qibin Liu$^{2}$, Xiaoxue Li$^{2}$, Zhuoyu Li$^{2}$, \\
Yangyu Tao$^{2}$, Fei Gao$^{2}$, Cheng Jiang$^{2}$, Bo Chao Wang$^{2}$, Kai Liu$^{2}$, Jianchen Zhu$^{2}$, \\
Wai Lam$^{3}$, Bo Zhou$^{1,}$\thanks{\ Project Lead.}, Di Wang$^{1}$\\
\textbf{$^1$LLM Department, Tencent} \quad \textbf{$^2$HunYuan Infra Team}\\
\textbf{$^3$The Chinese University of Hong Kong} \\
\Letter~chaysezhou@tencent.com\\
}

\begin{document}
\maketitle
\let\oldthefootnote\thefootnote

\let\thefootnote\oldthefootnote

\begin{abstract}

The growing disparity between the exponential scaling of computational resources and the finite growth of high-quality text data now constrains conventional scaling approaches for large language models (LLMs). To address this challenge, we introduce Reinforcement Learning on Pre-Training data (\ours), a new training-time scaling paradigm for optimizing LLMs. In contrast to prior approaches that scale training primarily through supervised learning, \ours\ enables the policy to autonomously explore meaningful trajectories to learn from pre-training data and improve its capability through reinforcement learning (RL). While existing RL strategies such as reinforcement learning from human feedback (RLHF) and reinforcement learning with verifiable rewards (RLVR) rely on human annotation for reward construction, \ours\ eliminates this dependency by deriving reward signals directly from pre-training data. Specifically, it adopts a next-segment reasoning objective, rewarding the policy for accurately predicting subsequent text segments conditioned on the preceding context. This formulation allows RL to be scaled on pre-training data, encouraging the exploration of richer trajectories across broader contexts and thereby fostering more generalizable reasoning skills. Extensive experiments on both general-domain and mathematical reasoning benchmarks across multiple models validate the effectiveness of \ours. For example, when applied to Qwen3-4B-Base, \ours\ yields absolute improvements of $3.0$, $5.1$, $8.1$, $6.0$, $6.6$, and $5.3$ on MMLU, MMLU-Pro, GPQA-Diamond, KOR-Bench, AIME24, and AIME25, respectively. The results further demonstrate favorable scaling behavior, suggesting strong potential for continued gains with more compute. In addition, \ours\ provides a solid foundation, extending the reasoning boundaries of LLMs and enhancing RLVR performance.

\end{abstract}
\begin{figure*}[h]
\centering
\includegraphics[width=1.0\linewidth]{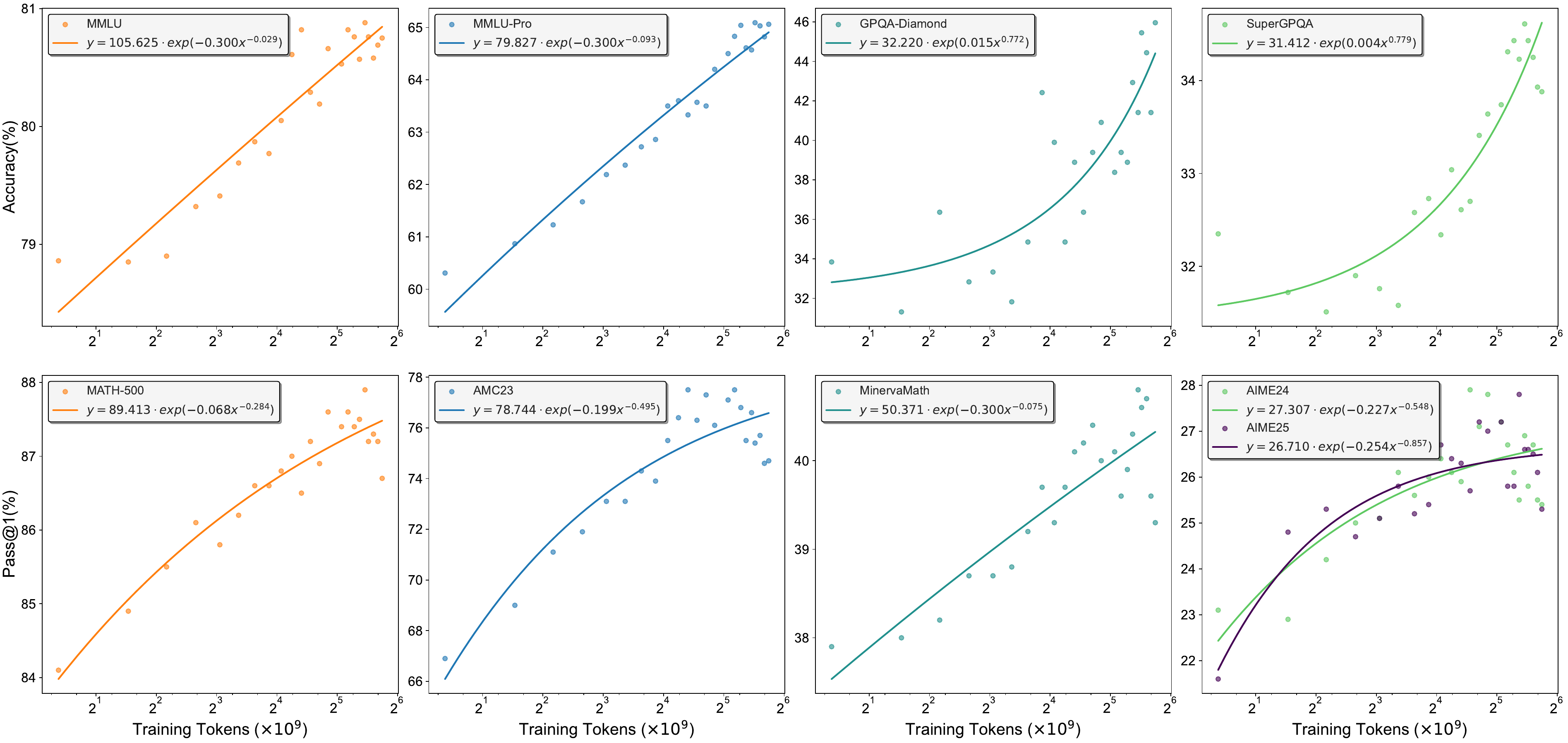}
\caption{Scaling law of \ours\ performance on various benchmarks with respect to training tokens.}
\label{fig:rpt_scaling}
\end{figure*}

\section{Introduction}

Large language models (LLMs) have achieved remarkable success across diverse domains, including human-aligned conversational assistants \citep{bai2022training, ouyang2022training} and autonomous AI agents \citep{team2025kimi}. A central driver of this progress has been the scaling of computational resources during training, realized through the simultaneous expansion of both data and model parameters. For instance, training corpora have grown from billions of tokens in BERT \citep{devlin2019bert} to trillions in Llama \citep{touvron2023llama, grattafiori2024llama}, while model sizes have scaled from millions of parameters in BERT \citep{devlin2019bert} to the trillion-parameter level in Kimi K2 \citep{team2025kimi}. However, parameter scaling requires increasingly demanding infrastructure and results in prohibitive inference costs, whereas data scaling is constrained by the scarcity of high-quality web corpora \citep{villalobos2024position, muennighoff2023scaling, ruan2025reasoning}.

In this paper, we propose a new scaling paradigm \ours\footnote{RLPT stands for Reinforcement Learning on Pre-Training data.} to optimize LLMs through reinforcement learning (RL) on pre-training data. In contrast to prior scaling approaches that primarily rely on supervised learning, \ours\ allocates training compute to enable the policy to autonomously explore meaningful reasoning trajectories to learn from pre-training data and improve its overall capabilities through reinforcement learning (RL). This paradigm offers two main advantages. First, it enables reasoning for learning: rather than directly learning token by token, the model generates intermediate reasoning content that can uncover the latent thought process underlying data construction, augment the original data, and support more data-efficient learning \citep{ruan2025reasoning}. Second, RL leverages self-explored trajectories for training, maintains proximity to the original policy distribution, and thereby fosters stronger generalization capabilities \citep{chu2025sft, lai2025reinforcement, shenfeld2025rl}. However, directly scaling RL also introduces new challenges, since existing frameworks such as reinforcement learning from human feedback (RLHF) \citep{bai2022training} and reinforcement learning with verifiable rewards (RLVR) \citep{guo2025deepseek} still rely heavily on human annotation, which constrains their scalability on pre-training data.


To address this challenge, we propose a novel next-segment reasoning objective that can obtaining meaningful self-supervised reward from unlabeled internat data. To be more specifically, the model is first required to predict the subsequent segment of text, and then the reward signal is derived by evaluating the semantic consistency between the predicted segment and the real segment using a generative reward model. Based on different prediction segment configurations, we propose two tasks with distinct effect. The first requires the model to predict a complete subsequent sentence given the preceding context, which we term the Autoregressive Segment Reasoning (ASR) task. The second involves a context with masked tokens in the middle, where the model must leverage both preceding and following context to infer a continuous span of masked tokens, which we designate as the Middle Segment Reasoning (MSR) task. During training, we interleave
ASR and MSR task to simultaneously optimize the model's autoregressive generation capabilities as well as the in-context understanding abilities.

We evaluate \ours\ across both general-domain and mathematical reasoning tasks using multiple models. Experimental results demonstrate that \ours\ delivers consistent and substantial improvements in both settings. For example, when applied to Qwen3-4B-Base, \ours\ achieves absolute gains of $3.0$, $5.1$, $8.1$, and $6.0$ on MMLU \citep{hendrycks2021measuringmmlu}, MMLU-Pro \citep{wang2024mmlu}, GPQA-Diamond \citep{rein2024gpqa}, and KOR-Bench \citep{ma2024kor}, respectively, together with improvements of $6.6$ and $5.3$ in Pass@$1$ on AIME24 and AIME25 \citep{aime}. Comparable gains are also observed on Llama3.2-3B-Base and Qwen3-8B-Base, with detailed results provided in Sec.~\ref{sec:main_results}. Beyond standalone performance, \ours\ also strengthens the reasoning capability of LLMs. When serving as the foundation for RLVR, it yields additional improvements of $2.3$ and $1.3$ in Pass@$1$, and $3.7$ and $2.0$ in Pass@$8$, on AIME24 and AIME25 with Qwen3-4B-Base, respectively. We further analyze the scaling behavior of \ours, showing that downstream performance empirically follows a scaling law with training compute (Fig.~\ref{fig:rpt_scaling}), highlighting its potential for continued progress with increased compute. In addition to quantitative results, qualitative analysis of reasoning trajectories reveals diverse reasoning strategies, providing insight into the effectiveness of \ours. Finally, we distill practical design lessons from \ours\ to inform future research in this direction.

Our contributions can be summarized in three aspects:
\begin{itemize}[leftmargin=1.6em]
    \item We propose RLPT, a method that scales RL on pre-training data. To remove the reliance on human annotation, we design a next-segment reasoning objective, consisting of ASR and MSR tasks, which reward LLMs for correctly predicting the ground-truth next segment given the preceding context.
    \item Extensive experiments on general-domain and mathematical reasoning tasks across multiple models show that RLPT substantially improves performance and exhibits a favorable scaling trend, empirically establishing a scaling law in benchmark performance as compute increases, indicating strong potential for continued gains.
    \item Results further demonstrate that RLPT provides a strong foundation for subsequent RLVR, extending the reasoning boundaries of LLMs and boosting performance on mathematical reasoning benchmarks.
\end{itemize}

\section{Preliminary}
In this section, we briefly review reinforcement learning (RL) and the supervised learning paradigm of next-token prediction in training large language models (LLMs), as these serve as the foundation of \ours.

\subsection{Reinforcement Learning in Large Language Models}
Reinforcement learning (RL) has become an essential component for improving LLMs. Formally,
\begin{equation}
    \mathcal{J}_{\text{RL}}(\theta) = \mathbb{E}{[q \sim D_{q}, o \sim \pi_{\theta}(\cdot\mid q)]} [r(o)],
\label{eq:rl}
\end{equation}
where $r(o)$ denotes the reward assigned to output $o$. In reinforcement learning from human feedback (RLHF) \citep{ouyang2022training, bai2022training}, rewards are provided by a neural reward model trained on human-annotated preference pairs. More recently, reinforcement learning with verifiable rewards (RLVR) employs rule-based functions that compare model outputs against reference answers \citep{guo2025deepseek, zeng2025simplerl}. Optimizing Eq.~\ref{eq:rl} encourages the model to reinforce behaviors associated with higher rewards while suppressing those linked to lower rewards. In practice, this objective is typically optimized using policy gradient algorithms such as PPO \citep{schulman2017proximal} and GRPO \citep{shao2024deepseekmath}. Despite their effectiveness, both RLHF and RLVR face scalability challenges due to their reliance on human annotations.

\subsection{Next-Token Prediction}
Next-token prediction (NTP) is the fundamental training objective of modern LLMs. Formally,
\begin{equation}
    \mathcal{J}_{\text{NTP}}(\theta) = \mathbb{E}{[x \sim \mathcal{D}_{x}]} - \frac{1}{|x|} \sum_{i=1}^{|x|} \log \pi_\theta(x_i \mid x_{<i}),
\end{equation}
where $x$ is a token sequence and $|x|$ its length. Pre-training and post-training based on NTP constitute the mainstream optimization paradigm for LLMs, yielding remarkable success across diverse applications. Nevertheless, recent studies suggest that supervised fine-tuning (SFT) under the NTP paradigm often promotes surface-level memorization rather than fostering the deeper generalization capabilities achievable with RL \citep{chu2025sft, lai2025reinforcement, shenfeld2025rl}.
\section{Reinforcement Learning on Pre-Training Data}
\begin{figure*}[t]
\centering
\includegraphics[width=1.0\linewidth]{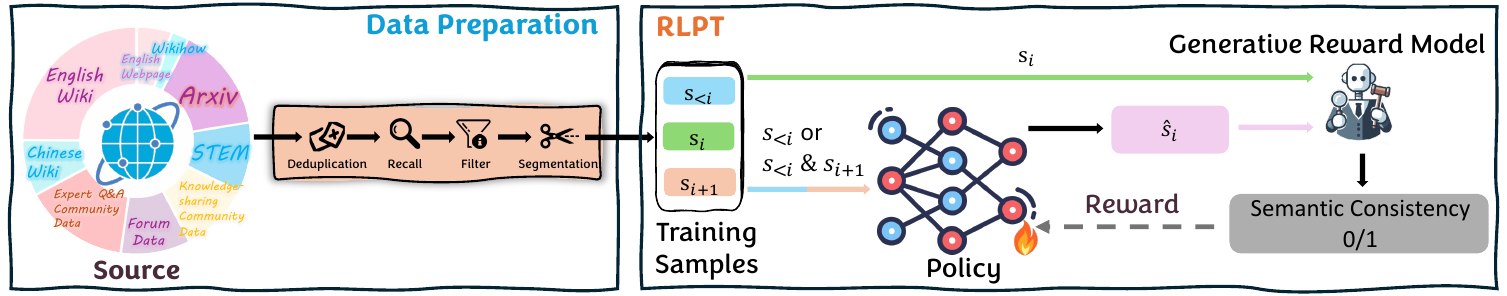}
\caption{Overview of \ours. Raw data from internet corpora is processed into training samples of the form $(s_{<i}, s_{i}, s_{i+1})$. During the reinforcement pre-training stage, the policy LLM predicts $\hat{s}_{i}$ conditioned on $s_{<i}$ (ASR) or on $(s_{<i}, s_{i+1})$ (MSR). The prediction is then compared with $s_{i}$ to compute the reward.}
\label{fig:method_overview}
\end{figure*}

To address the limitations of scalability and generalization, we propose Reinforcement Learning on Pre-Training data (\ours). In this framework, next-segment reasoning serves as the reinforcement learning (RL) objective, where the subsequent segment in unlabeled text acts as the ground truth. This self-supervised objective removes the reliance on human annotation and enables RL to scale directly on large pre-training corpora. An overview of \ours\ is shown in Fig. \ref{fig:method_overview}.

\subsection{Data Preparation}
We construct a corpus for \ours\ by aggregating web text from diverse sources such as Wikipedia, arXiv, and threaded conversation data. To ensure data quality and compliance, we apply a multi-stage preprocessing pipeline consisting of: (i) MinHash-based near-deduplication, (ii) detection and masking of personally identifiable information (PII), and (iii) contamination removal with respect to all development and evaluation sets. Given the inherent noise in web corpora, we further implement a rigorous filtering procedure that integrates both rule-based and model-based methods. The rule-based stage eliminates content that is clearly unsuitable for language model training, whereas the model-based stage employs an instruction-tuned language model to perform more fine-grained quality assessments. Furthermore, we curated high-quality QA data from the annealing dataset \cite{team2025hunyuan} for mathematical reasoning tasks to enhance the model's reasoning ability.

\subsection{Next-Segment Reasoning}
Given a text $t$ from the pre-training data, we divide it into a sequence of contiguous segments $t = [s_1, s_2, \ldots, s_n]$, where each $s_i$ corresponds to a semantically coherent unit such as a phrase, a complete sentence, or a reasoning step. We then construct a dataset
\begin{equation}
\mathcal{D}s = {(s_{<i}, , s_i, s_{i+1}) \mid i = 2, \dots, n-1 },
\end{equation}
where $s_{<i} = [s_1, s_2, \dots, s_{i-1}]$ denotes the context, $s_i$ is the target segment, and $s_{i+1}$ is its subsequent segment. Based on this formulation, we introduce two segment-level training objectives that capture richer semantics than token-level prediction. Inspired by next-token prediction (NTP), we propose Autoregressive Segment Reasoning (ASR), which trains the policy to predict $s_i$ from $s_{<i}$, aligning with the autoregressive generation process of modern LLMs. To further enable the model to leverage broader contextual information, we introduce Middle Segment Reasoning (MSR), which trains the policy to predict $s_i$ from both $s_{<i}$ and $s_{i+1}$. This resembles masked language modeling \citep{devlin2019bert, liu2019roberta, raffel2020exploring} and is particularly useful for tasks such as code completion. During training, we interleave ASR and MSR by designing different prompts and extracting the predicted segment between special tags in the output. The prompt for the ASR task is illustrated below.
\begin{prompt}[notitle]{-15pt}{-5pt}{}
    Complete the text provided under ### Context by predicting the next most probable sentence. 
    
    Please reason step by step to determine the best possible continuation, and then enclose your final answer within <|startofprediction|> and <|endofprediction|> tags.
    
    ### Context
    
    {context} 
\end{prompt}

Similarly, the prompt for the MSR task is presented as follows
\begin{prompt}[notitle]{-15pt}{-5pt}{}
    ## Text Material ##:
    {prompt}
    
    <MASK>

    {next_step}
    
    ## Task ##: 
    Fill in the <MASK> section of the material with appropriate sentences or a solution step.
    
    Carefully reason step by step to determine the most suitable completion.  
    Finally, provide your best prediction for the <MASK> section.  
    Enclose your final answer for the <MASK> part within <|startofprediction|> and <|endofprediction|>.
\end{prompt}

The reward is defined as the semantic consistency between the predicted and reference segments, evaluated by a generative reward model $G_{rm}$. This model assesses whether the two segments convey equivalent content while allowing for linguistic variation. In practice, we find that directly comparing the predicted segment with the ground-truth next segment is overly strict, since the model may generate outputs that span multiple subsequent segments. To address this issue, we provide $G_{rm}$ with several subsequent segments as reference and instruct it to verify whether the predicted segment is a valid prefix of the reference content. The prompt for $G_{rm}$ is shown below.
\begin{prompt}[notitle]{-15pt}{-5pt}{}
    ## Task
    Given a Predicted sentence and a Reference paragraph, determine whether the Predicted text is a prefix (initial segment) of the Reference paragraph, and whether it expresses exactly the same semantic content as the corresponding prefix of the Reference. 
    The Predicted text does not need to match the prefix of the Reference word-for-word, but it must convey the same meaning.
    
    Reference:
    {reference}
    
    Predicted:
    {predicted}
    
    ## Scoring Rules
    
    If the Predicted text semantically matches the prefix of the Reference, assign a score of 1.
    If the Predicted text does not semantically match the prefix of the Reference, assign a score of 0.
    When making your judgment, focus primarily on semantic equivalence, not on exact wording.
    
    Only output the score on a single line; do not provide any explanatory text or additional content. 
    Output format (choose one):
    
    Score: 0  
    or  
    Score: 1
\end{prompt}

Given a predicted segment $\hat{s}_{i}$ extracted from the model output $o$, the reward is specified as
\begin{equation}
r(o, s_{i}) =
\begin{cases}
1 & \text{if } G_{rm}(\hat{s}_{i}, s_{i}) = 1, \\
0 & \text{otherwise}.
\end{cases}
\end{equation}

The training objective of \ours\ is defined as
\begin{equation}
\begin{aligned}
    \mathcal{J}_{\text{SRPT}}(\theta) 
    &= \mathbb{E}_{ASR}{[(s_{<i},s_{i}) \sim \mathcal{D}_{s}, \, o \sim \pi_{\theta}(\cdot \mid s_{< i})]} [r(o,s_{i})] \\
    &\quad + \lambda \, \mathbb{E}_{MSR}{[(s_{<i},s_{i}, s_{i+1}) \sim \mathcal{D}_{s}, \, o \sim \pi_{\theta}(\cdot \mid s_{< i}, s_{i+1})]} [r(o,s_{i})],
\end{aligned}
\end{equation}
where $\lambda \in (0, 1)$ is a hyperparameter that balances the contributions of ASR and MSR terms, and may be adjusted depending on the requirements of specific downstream applications.

\subsection{Training Details}
\paragraph{Cold-Start.}
\ours\ can be applied to a base model after next-token pre-training, but it requires a minimum level of instruction-following ability to initiate next-segment reasoning. To satisfy this requirement, we introduce a cold-start phase consisting of supervised fine-tuning on instruction-following data.

\paragraph{Next-Segment Reasoning.}
In this work, we define a segment unit as a sentence by default. We also conducted preliminary studies with alternative segmentation units, such as employing LLMs to extract integrated atomic steps from text, but these approaches did not yield clear improvements over sentence-level segmentation. Therefore, we adopt sentence segmentation as the default setting in our experiments and leave the exploration of other strategies for future work. For sentence segmentation, we use the NLTK toolkit \citep{bird2006nltk}, filtering out sentences that are too short. Each remaining sentence is then treated as a target for RL under the next-segment reasoning objective.
\section{Experiments}
\subsection{Experimental Setup}
Experiments are conducted on Llama3 models \citep{grattafiori2024llama} and Qwen3 models \citep{yang2025qwen3}. In the cold-start supervised fine-tuning (SFT) stage, we use a batch size of $1024$, a learning rate of $2 \times 10^{-5}$ with a cosine scheduler, and train for $3$ epochs. For next-segment reasoning, we adopt a batch size of $512$, a maximum response length of $8192$, and a constant learning rate of $1 \times 10^{-6}$. For each prompt, we sample $8$ outputs with a temperature of $1.0$, and optimization is performed using on-policy GRPO \citep{shao2024deepseekmath} without KL regularization. In the mathematical reasoning domain, we further conduct RLVR experiments, evaluating its performance when built on \ours, with RLVR configured using the same hyperparameters as the next-segment reasoning task.

\subsection{Evaluation Metric}
We evaluate model performance on both general-domain and mathematical reasoning tasks. For the general domain, we use benchmarks including MMLU \citep{hendrycks2021measuringmmlu}, MMLU-Pro \citep{wang2024mmlu}, GPQA-Diamond \citep{rein2024gpqa}, SuperGPQA \citep{du2025supergpqa}, KOR-Bench \citep{ma2024kor}, and OlympiadBench \citep{he2024olympiadbench}, reporting accuracy as the evaluation metric. For mathematical reasoning, we evaluate on MATH-500 \citep{hendrycks2021measuring}, AMC23 \citep{amc}, Minerva Math \citep{lewkowycz2022solving}, and AIME \citep{aime}, using the Pass@$k$ metric, which measures the probability that at least one correct solution appears among $k$ independent attempts. We adopt the unbiased estimator of Pass@$k$ \citep{chen2021evaluating}:
\begin{equation}
\text{Pass@}k = \mathbb{E}_{\boldsymbol{x} \sim \mathcal{D}} \left[1 - \frac{\binom{n - c}{k}}{\binom{n}{k}} \right],
\end{equation}
where $n$ is the number of sampled responses per prompt and $c$ is the number of correct responses. We sample $n=64$ responses with temperature $0.6$ and top-$p$ $0.95$, and report Pass@$1$ and Pass@$8$. The maximum generation length is set to $32{,}768$ tokens. Correctness in mathematical reasoning is evaluated using Math-Verify\footnote{\url{https://github.com/huggingface/Math-Verify}}.

\subsection{Experimental Results}
\label{sec:main_results}
\begin{table}[ht]
  \centering
  \vspace{1ex}
  \renewcommand{\arraystretch}{1.0}
  \resizebox{0.95\linewidth}{!}{%
    \begin{tabular}{lcccccc}
      \toprule
      \textbf{Training} 
        & \textbf{MMLU} & \textbf{MMLU-Pro} & \textbf{GPQA-Diamond}  & \textbf{SuperGPQA} & \textbf{KOR-Bench} & \textbf{OlympiadBench} \\
      \midrule
      \midrule
      \multicolumn{7}{c}{\textit{Llama-3.2-3B-Base}} \\
      \midrule
      \midrule
      Base                  & $4.2$            & $21.3$            & $3.5$            & $7.7$            & $3.1$             & $1.7$  \\
      \midrule
      \quad $+$ Cold-Start  & $59.4$            & $34.7$            & $16.7$            & $15.8$            & $39.1$            & $14.4$  \\
      \quad $+$ \ours       & $59.4$ & $\mathbf{36.2}$     & $\mathbf{28.3}$   & $\mathbf{19.2}$   & $\mathbf{39.4}$   & $\mathbf{15.9}$\\
      \midrule
      \midrule
      \multicolumn{7}{c}{\textit{Qwen3-4B-Base}} \\
      \midrule
      \midrule
      Base                  & $30.6$            & $16.0$            & $17.7$            & $25.4$            & $3.7$             & $35.0$  \\
      \midrule
      \quad $+$ Cold-Start  & $77.8$            & $59.7$            & $31.3$            & $32.3$            & $50.7$            & $51.7$  \\
      \quad $+$ \ours       & $\mathbf{80.8}$ & $\mathbf{64.8}$     & $\mathbf{39.4}$   & $\mathbf{34.3}$   & $\mathbf{56.7}$   & $\mathbf{52.7}$\\
      \midrule
      \midrule
      \multicolumn{7}{c}{\textit{Qwen3-8B-Base}} \\
      \midrule
      \midrule
      Base                  & $58.9$            & $47.0$            & $27.8$            & $28.5$            & $40.6$             & $38.8$  \\
      \midrule
      \quad $+$ Cold-Start  & $81.6$            & $64.9$            & $45.5$            & $37.8$            & $55.1$            & $57.6$  \\
      \quad $+$ \ours       & $\mathbf{83.0}$ & $\mathbf{68.3}$     & $\mathbf{47.5}$   & $\mathbf{40.1}$   & $\mathbf{55.7}$   & $\mathbf{59.7}$\\
      \bottomrule
    \end{tabular}%
  }
  \vspace{1ex}
  \caption{performance on general-domain tasks across different models, with the best results highlighted.}
  \label{tab:general_results}
\end{table}
\begin{table}[ht]
  \centering
  \vspace{1ex}
  \renewcommand{\arraystretch}{1.1}
  \resizebox{1.0\linewidth}{!}{%
    \begin{tabular}{lcccccccccc}
      \toprule
      \multirow{2}{*}{\textbf{Training}} 
        & \multicolumn{5}{c}{\textbf{Pass@$1$}} 
        & \multicolumn{5}{c}{\textbf{Pass@$8$}} \\
      \cmidrule(lr){2-6} \cmidrule(lr){7-11}
      & \textbf{MATH} & \textbf{AMC23}  & \textbf{Minerva} & \textbf{AIME24} & \textbf{AIME25}
      & \textbf{MATH} & \textbf{AMC23}  & \textbf{Minerva} & \textbf{AIME24} & \textbf{AIME25} \\
      \midrule
      Base         & $39.8$        & $24.7$        & $17.7$        & $7.3$        & $4.5$    
                   & $79.9$        & $65.6$        & $41.0$        & $24.3$       & $21.6$        \\
      \midrule
      \quad $+$ Cold-Start & $83.6$        & $65.9$        & $38.2$        & $20.6$        & $21.9$    
                           & $95.0$        & $91.8$        & $54.1$        & $40.3$        & $39.5$        \\
      \quad $+$ \ours  & $87.4$        & $77.1$        & $40.1$        & $27.2$        & $27.2$    
                       & $95.3$        & $92.1$        & $54.8$        & $45.3$        & $40.9$        \\
      \midrule
      \quad $+$ RLVR   & $89.1$        & $76.3$        & $41.6$        & $27.6$        & $27.7$    
                       & $96.7$        & $\mathbf{94.3}$        & $55.8$        & $49.8$        & $41.6$ \\
      \quad $+$ \ours $+$ RLVR   
                       & $\mathbf{90.6}$ & $\mathbf{79.7}$ & $\mathbf{42.1}$ & $\mathbf{29.9}$ & $\mathbf{29.0}$    
                       & $\mathbf{96.8}$ & $93.5$ & $\mathbf{56.8}$ & $\mathbf{53.5}$ & $\mathbf{43.6}$        \\
      \bottomrule
    \end{tabular}%
  }
  \vspace{1ex}
  \caption{Performance on mathematical reasoning benchmarks based on the Qwen3-4B-Base model with $64$ samples per prompt, the best performance are highlighted.}
  \label{tab:math_results}
\end{table}

\paragraph{General Domain.}
The performance on general-domain tasks is summarized in Tab.~\ref{tab:general_results}, where \ours\ delivers substantial and consistent gains across all benchmarks and models. In particular, when applied to Qwen3-4B-Base, it achieves absolute improvements of $3.0$, $5.1$, $8.1$, $2.0$, and $6.0$ on MMLU, MMLU-Pro, GPQA-Diamond, SuperGPQA, and KOR-Bench, respectively. With Qwen3-8B-Base, the improvements are $1.4$, $3.4$, $2.0$, $2.3$, and $2.1$ on MMLU, MMLU-Pro, GPQA-Diamond, SuperGPQA, and OlympiadBench, respectively. Furthermore, results on Llama-3.2-3B-Base confirm the generalizability of \ours\ across different model families, with absolute improvements of $1.5$, $11.6$, and $3.4$ on MMLU-Pro, GPQA-Diamond, and SuperGPQA, respectively. Since these benchmarks span diverse domains including STEM, law, economics, and health, the results demonstrate that \ours\ effectively leverages the extensive knowledge contained in large-scale pre-training corpora.

\paragraph{Mathematical Reasoning.}
As shown in Tab.~\ref{tab:math_results}, \ours\ yields substantial gains in mathematical reasoning, improving performance on both Pass@$1$ and Pass@$8$. On the challenging AIME24 and AIME25 benchmarks, \ours\ achieves absolute improvements of $6.6$ and $5.3$ on Pass@$1$, and $5.0$ and $1.4$ on Pass@$8$, respectively. These improvements indicate that \ours\ is effective in unlocking the reasoning boundary, thereby providing a strong basis for subsequent RLVR training. Indeed, when \ours\ is used as the initialization for RLVR, it further boosts performance, with absolute gains of $2.3$ (AIME24) and $1.3$ (AIME25) on Pass@$1$, and $3.7$ (AIME24) and $2.0$ (AIME25) on Pass@$8$. This demonstrates that \ours\ enhances both exploitation and exploration, which are typically considered competing objectives.

\begin{figure*}[h]
\centering
\includegraphics[width=1.0\linewidth]{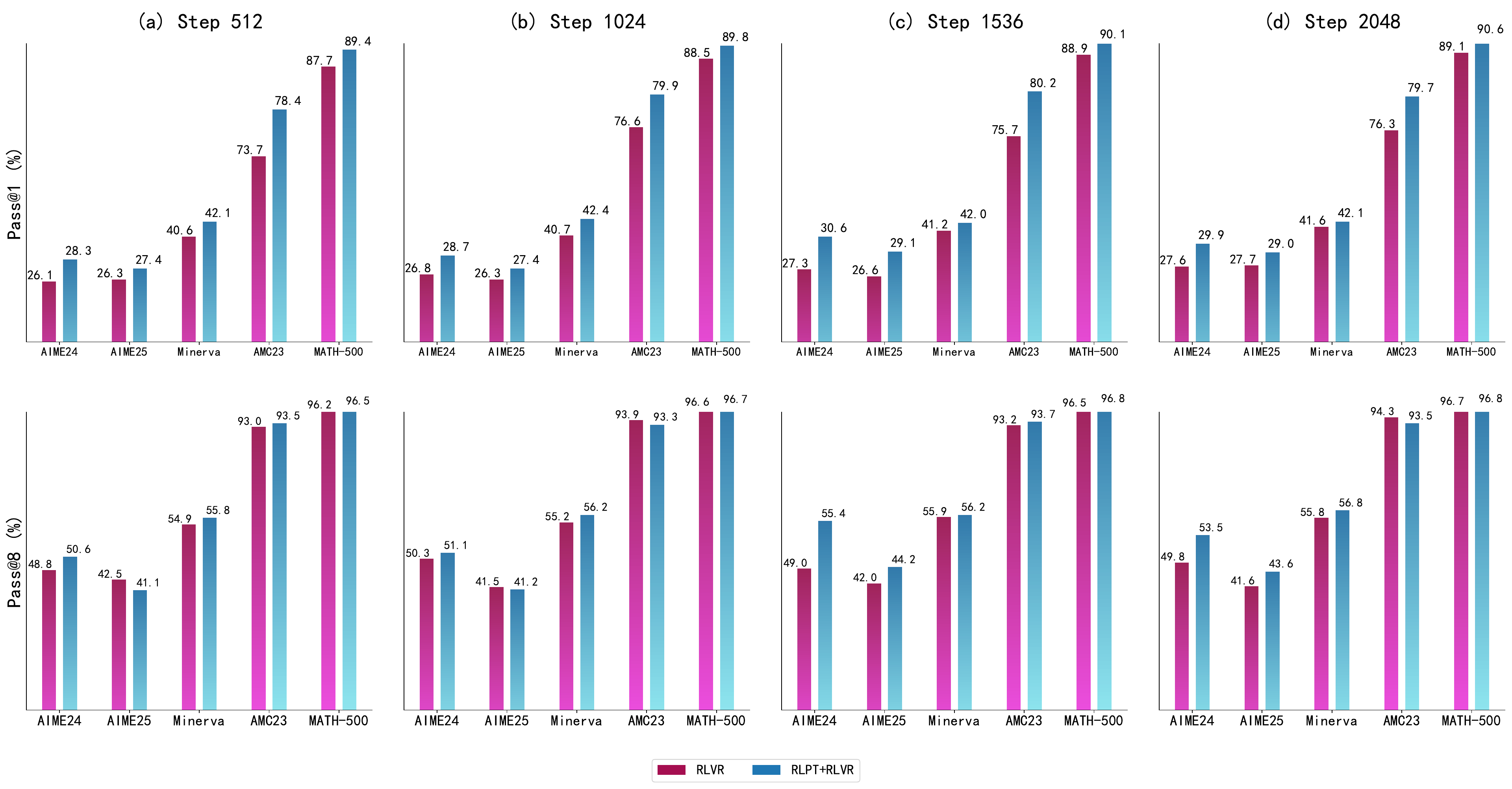}
\caption{Comparative scaling properties of RLVR and \ours\ $+$ RLVR.}
\label{fig:rpt_rlvr_scaling}
\end{figure*}

\subsection{Analysis}

\paragraph{Scaling Properties.}
As shown in Fig. \ref{fig:rpt_scaling}, the performance of \ours\ on various benchmarks follows a power-law decay with respect to the number of training tokens, suggesting potential for further gains through scaling compute. We also report the scaling trend when \ours\ serves as the foundation for RLVR in Fig. \ref{fig:rpt_rlvr_scaling}. In this setting, \ours\ provides a strong initialization, yielding consistent improvements throughout training. Notably, both Pass@$1$ and Pass@$8$ improve, indicating that the gains from \ours\ do not come at the expense of exploration capability, which remains a common challenge for current RLVR methods \citep{yue2025does, walder2025pass, chen2025pass, wu2025invisible}.

\paragraph{Thinking Patterns.}
\begin{table}[t]
\centering
\scalebox{1.0}{
\begin{tabular}{p{15cm}}
\toprule
\textbf{Thinking Process of \ours} \\
\toprule
Alright, let's tackle this step by step. I'm going to predict the next most probable sentence after the given context... \newline
\textcolor{myred}{Understanding the Context:} the Fourier transform of $\chi_{[-a,a]}$ is defined, the integral is restricted to $[-a,a]$, the exponential is split into cosine and sine parts... The cosine integral gives $\tfrac{2 \sin(a \xi)}{\xi}$, while the sine integral gives $0$... \newline
\textcolor{myred}{Identifying the Next Step:} the next move would be to combine these results... Since the sine term vanishes, the final answer should come only from the cosine contribution... \newline
\textcolor{myblue}{Verifying the Flow:} after evaluating both parts, the solution normally concludes by combining them into the closed-form Fourier transform... \newline
\textcolor{myblue}{Considering Alternative Path:} could the solution instead highlight the behavior at $\xi=0$ or present a piecewise definition... \textcolor{myblue}{Alternatively,} the solution might emphasize this special case, yet it is usually handled by noting the limit, so the general form is sufficient... \newline
\textcolor{myblue}{Looking back,} one might also check the special case $\xi=0$, but the limit already gives $2a$, so the concise expression is consistent... \newline
\textcolor{myred}{So the natural continuation is:} \texttt{<|startofprediction|>}Thus, combining these results, the Fourier transform of $\chi_{[-a,a]}$ is given by $F(\xi) = \tfrac{2 \sin(a \xi)}{\xi}$. \texttt{<|endofprediction|>} \\
\bottomrule
\end{tabular}}
\caption{Example of reasoning patterns of \ours. Task-related keywords are highlighted in \textcolor{myblue}{blue}, while reasoning-related keywords are highlighted in \textcolor{myred}{red}.}
\label{tab:thinking_pattern} 
\end{table}
To demonstrate the benefits of \ours, we provide in Tab.~\ref{tab:thinking_pattern} an illustrative example of its reasoning process. In this case, the model approaches the next-segment reasoning task through a structured sequence: it first abstracts the preceding context to capture the overarching flow, then determines the subsequent step, formulates a candidate continuation, verifies its plausibility, explores alternative possibilities, performs backtracking when appropriate, and ultimately produces the final answer. This structured trajectory aligns with the multi-step reasoning strategies exhibited by LLMs in complex problem-solving \citep{guo2025deepseek, jaech2024openai}, which helps explain the effectiveness of \ours.

\paragraph{Reward Modeling.}

\begin{figure*}
\centering
\includegraphics[width=1.0\linewidth]{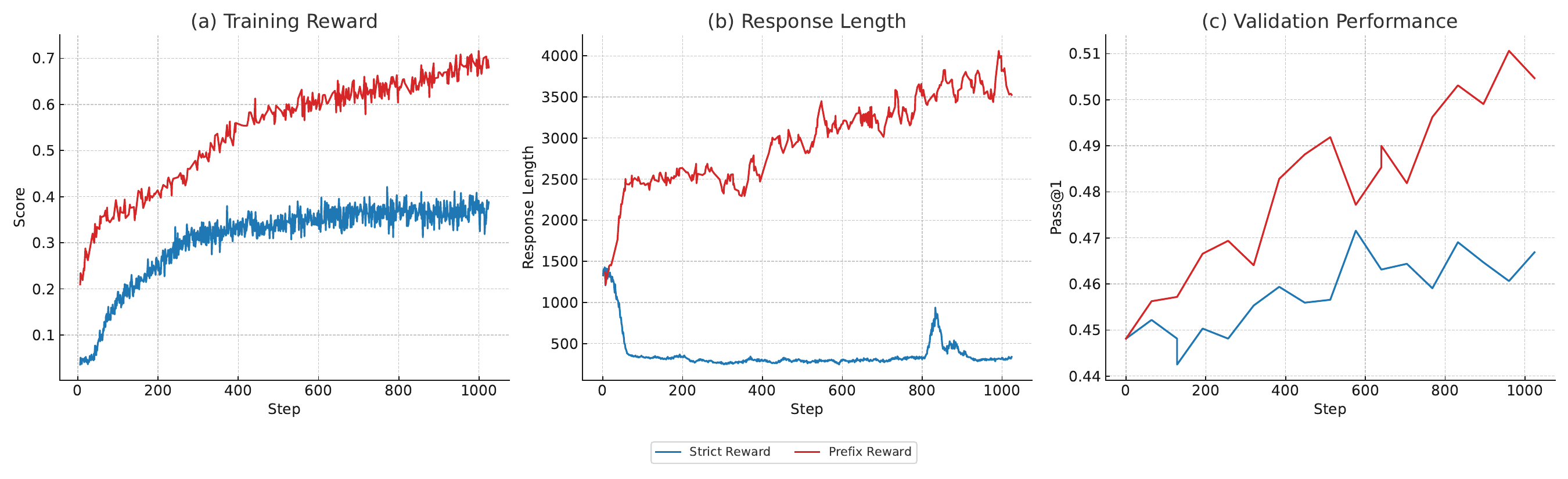}
\caption{Comparison between Strict Reward and Prefix Reward: (a) Training Reward, (b) Response Length, (c) Validation Performance (Pass@$1$).}
\label{fig:reward_modeling}
\end{figure*}

In developing \ours, we iteratively refined our reward modeling approach after encountering several challenges with our initial formulation.  Our initial approach adopted a strict reward that required the predicted segment to convey exactly the same semantic content as the ground-truth segment. This constraint proved too rigid, leading to numerous false positives. We observed that the model often generated outputs that encompassed multiple ground-truth segments, largely due to the uneven distribution of information across sentence-based segmentation: some sentences contained only a single formula, while others might captured the complete solution to a subproblem. Such discrepancies disrupted the training process and yielded only limited improvements in downstream performance, as illustrated in Fig.~\ref{fig:reward_modeling}. To address this issue, we introduce a relaxed prefix reward, which assigns a score of $1$ as long as the predicted segment forms a valid prefix of the ground-truth completion. This adjustment addresses segments with varying information content and provides a more stable training signal. It also enables the model to generate longer responses, which in turn results in improved performance on downstream mathematical reasoning tasks, as shown in Fig.~\ref{fig:reward_modeling}.
\section{Related Work}

\paragraph{Scaling Paradigms.}
The progress of language models has been fundamentally driven by scaling compute, which can be broadly divided into training-time scaling and test-time scaling. Training-time scaling primarily relies on next-token prediction, increasing computational cost by enlarging model size or expanding pre-training data to reduce prediction loss \citep{radford2019language, brown2020language, kaplan2020scaling, hoffmann2022training}. In contrast, test-time scaling allocates more compute during inference by generating extended chains of reasoning before producing the final answer \citep{brown2024large, jaech2024openai, muennighoff2025s1, guo2025deepseek}. \ours\ belongs to the training-time scaling paradigm but differs from prior approaches that emphasize supervised learning. Instead, it employs reinforcement learning (RL), allocating compute for the model to self-explore and learn from large-scale pre-training corpora. RL provides two notable advantages. First, it enables the model to uncover the latent reasoning underlying data, which can be regarded as a compressed form of deliberative thinking reflected in scientific papers or textbooks \citep{ruan2025reasoning}. Second, recent research suggests that RL supports better generalization compared with supervised learning \citep{chu2025sft, lai2025reinforcement, shenfeld2025rl}. The most relevant approaches are RPT \citep{dong2025reinforcement} and Quiet-STaR \citep{zelikman2024quietstar}, both of which apply RL on unlabeled data for training-time scaling. However, \ours\ differs by focusing on next-segment prediction rather than next-token prediction.

\paragraph{Reinforcement Learning in LLMs.}
RL has become a central paradigm for LLMs. Early applications mainly focused on aligning model outputs with human values \citep{bai2022training, ouyang2022training, mu2024rule}, typically through reward models trained on human-annotated preference pairs. More recently, RL has been used to strengthen reasoning abilities by leveraging rule-based reward functions that evaluate outputs against reference answers \citep{guo2025deepseek, zhu2025surprising}. Despite these advances, both directions ultimately depend on human-provided or verifiable supervision, which limits scalability. In contrast, \ours\ introduces the next-segment reasoning objective, where the subsequent segment in natural text serves as the reference. This design removes the need for human annotation and enables RL to scale effectively on large-scale pre-training data.

\section{Conclusion}
This work introduces \ours, a new training-time scaling paradigm that applies reinforcement learning to pre-training data. At its core, \ours\ adopts a self-supervised next-segment reasoning objective, which removes the need for human annotations and enables RL training on large unlabeled corpora. Extensive experiments demonstrate the effectiveness of \ours, yielding substantial gains in both general-domain and mathematical reasoning tasks. Moreover, the performance exhibits favorable scaling properties with respect to training compute, suggesting strong potential for further gains.

\bibliography{iclr2025_conference}
\bibliographystyle{iclr2025_conference}

\appendix

\end{document}